\documentclass{article}

\usepackage{PRIMEarxiv}

\usepackage[utf8]{inputenc} 
\usepackage[T1]{fontenc}    
\usepackage{hyperref}       
\usepackage{url}            
\usepackage{booktabs}       
\usepackage{amsfonts}       
\usepackage{nicefrac}       
\usepackage{microtype}      
\usepackage{lipsum}
\usepackage{fancyhdr}       
\usepackage{graphicx}       
\graphicspath{{media/}}     

\title{Approaching human 3D shape perception \\ with neurally mappable models}

\author{
  Thomas P. O'Connell \\
  Brain \& Cognitive Sciences \\
  MIT \\
  Cambridge, MA. USA\\
  \texttt{tpo@mit.edu} \\
  \\
  \And
  Tyler Bonnen \\
  EECS \\
  University of California, Berkeley \\
  Berkeley, CA. USA\\
  \AND
  Yoni Friedman \\
  Brain \& Cognitive Sciences \\
  MIT \\
  Cambridge, MA. USA \\
  \And
  Ayush Tewari \\
  CSAIL \\
  MIT \\
  Cambridge, MA. USA \\
  \And
  Josh B. Tenenbaum \\
  Brain \& Cognitive Sciences \\
  CSAIL \\
  MIT \\
  Cambridge, MA. USA \\
  \And
  Vincent Sitzmann \\
  CSAIL \\
  MIT \\
  Cambridge, MA. USA \\
  \And
  Nancy Kanwisher \\
  Brain \& Cognitive Sciences \\
  MIT \\
  Cambridge, MA. USA \\
}

\begin{document}
\maketitle

\begin{abstract}
Humans effortlessly infer the 3D shape of objects. What computations underlie this ability? Although various computational models have been proposed, none of them capture the human ability to match object shape across viewpoints. Here, we ask whether and how this gap might be closed. We begin with a relatively novel class of computational models, 3D neural fields, which encapsulate the basic principles of classic analysis-by-synthesis in a deep neural network (DNN). First, we find that a 3D Light Field Network (3D-LFN) supports 3D matching judgments well aligned to humans for within-category comparisons, adversarially-defined comparisons that accentuate the 3D failure cases of standard DNN models, and adversarially-defined comparisons for algorithmically generated shapes with no category structure. We then investigate the source of the 3D-LFN's ability to achieve human-aligned performance through a series of computational experiments. Exposure to multiple viewpoints of objects during training and a multi-view learning objective are the primary factors behind model-human alignment; even conventional DNN architectures come much closer to human behavior when trained with multi-view objectives. Finally, we find that while the models trained with multi-view learning objectives are able to partially generalize to new object categories, they fall short of human alignment. This work provides a foundation for understanding human shape inferences within neurally mappable computational architectures.
\end{abstract}

\section{Introduction}
When we look at an object, we do more than simply recognize its category and position in space. We also perceive the orientation and curvature of the visible surfaces that define the object, as well as its global 3D shape. We can imagine what the object would look like if we were to move around it. We retrieve action affordances such as how we would grasp or interact with the object. 

How do humans infer the 3D structure of objects? Several computational accounts have been proposed over the years, including analysis-by-synthesis approaches which invert generative shape models \cite{yuillek06, yildirim2020efficient} and more recently deep neural network models (DNNs) including convolutional neural networks (CNNs) and vision transformers (ViTs) \cite{kriegeskorte2015deep, yamins2016using, doerig2023neuroconnectionist}. However, no computational model has been proposed that performs human-level 3D shape inferences or scales to large naturalistic shape spaces. The human ability to perceive the 3D shape of objects thus remains one of the fundamental unsolved problems in vision.

Over the past 10 years, advances in DNNs have provided a new methodological framework for studying the primate visual system. Early progress was driven by convolutional neural networks (CNNs) trained for object classification on large-scale datasets (e.g., ImageNet). These models are appealing, in part because their architecture was loosely inspired by the organization of the primate visual system and their internal representations are neurally mappable to actual brain recordings. This modeling framework has enabled the prediction of behavioral and neural responses directly from images \cite{yamins2014performance, khaligh2014deep, rajalingham2018large, schrimpf2020integrative} and offers the most quantitatively accurate accounts of neural responses and behaviors that depend on the ventral visual stream \cite{kriegeskorte2015deep, yamins2016using, doerig2023neuroconnectionist}. 

Nonetheless, significant gaps remain between standard DNNs and human performance on many visual tasks, most notably the recovery of object shape from images. Unlike humans, ImageNet CNNs are biased to classify images based on texture \cite{geirhos2018imagenet, baker2018deep, hermann2020origins}. In 3D shape tasks that require matching objects across viewpoints, humans outperform standard CNNs by a wide margin \cite{rajalingham2018large, bonnen2021ventral}, an effect that holds even for the most recent CNN and ViT architectures (brainscore.org). Evidently, training models on large sets of natural scene images does not suffice for capturing human inferences of the 3D object shape.

This gap between human 3D shape perception and standard DNNs is well recognized in the computer vision community \cite{abbas2023progress, alcorn2019strike, geirhos2018imagenet, cooper2021out, reizenstein2021common}. Several computational methods have recently been developed to close this gap, with 3D neural fields driving many recent advances \cite{xie2022neural}. Unlike standard DNNs, these models were designed to explicitly account for the 3D properties of objects and scenes. 3D neural fields learn a continuous function that maps xyz or ray coordinates from a 3D volume to shape and/or color, given the objects pose. The most common class of 3D neural fields, NEural Radiance Fields (NERFs), are optimized directly on many views of an individual object or scene, and can be used to create near photo-realistic 3D models \cite{mildenhall2021nerf}. Other methods, such as conditional 3D neural fields, learn a generalizable shape space which can recover the global 3D shape of objects from a single image \cite{yu2021pixelnerf, sitzmann2021lfns}. Such conditional 3D neural fields better reflect what we expect from human shape perception: a system that can estimate the 3D shape of many objects when given an image from a single viewpoint.

3D neural fields also provide a bridge from classic approaches to model 3D shape perception and neurally mappable DNN models. 3D neural fields instantiate the basic principles of analysis-by-synthesis: inverting a generative model to infer a representation that can then be used to reproduce an image of an object or scene \cite{yuillek06}. Analysis-by-synthesis approaches to 3D vision have been considered for decades \cite{barrow1978recovering, lee2003hierarchical, blanz2023morphable, TejasDKulkarni:2015:cf27d, Martinez02}, but typically relied on top-down stochastic search algorithms that made them prohibitively slow. More recent work replaced stochastic top-down inference with fast amortized inference via a CNN to invert a generative face model \cite{yildirim2020efficient}, showing that the features learned in late layers of the inference CNN match the hierarchy of representations in the macaque face patch system and human behavior on 3D face matching tasks. This work, while only applied to a limited domain of visual stimuli, points to the promise of capturing generic 3D vision by combining DNNs with classic analysis-by-synthesis theories.

Here, we evaluate the alignment between 3D Light Field Networks (3D-LFNs) and human 3D shape judgements, with the aim of determining whether and how DNNs might achieve human-aligned performance on 3D shape matching tasks. First, we construct a 3D matching task in which human participants match images depicting the same object from two different viewpoints. As expected, we observe a large gap between standard DNNs and humans. Next, we build a 3D-LFN trained with a multi-view learning objective. We find that humans and the 3D-LFN make highly similar 3D shape inferences for within-category judgments on man-made objects. We create a series of ‘adversarial’ conditions based on 25 ImageNet CNNs, and find that the performance of humans and 3D-LFNs is robust to these adversarial trials, effects that generally hold for abstract procedurally-generated objects with no category structure. To determine what properties of the 3D-LFN lead to human-like 3D shape judgments, we conduct a series of computational experiments. Models trained on multiple viewpoints of the same objects with a learning objective that enforces a common 3D representation across viewpoints are best aligned to humans. Finally, we test the limitations of 3D-LFNs and other multi-view DNNs, finding that while they partially generalize to novel shape categories not included in their training diet, their alignment to human behavior drops markedly. Overall, we demonstrate that DNNs are capable of human-aligned 3D shape judgements when trained with exposure to multiple viewpoints of objects and the right multi-view learning objective, providing a framework for studying human 3D shape inferences in neurally-mappable architectures and identifying out-of-domain generalization to novel shape categories as a primary challenge for future research.

\section{Results}
\label{sec:results}

\subsection{3D shape judgements in humans and standard DNNs}

To probe 3D shape representations in humans and DNNs, we use a multi-view match-to-sample task. For the human experiments, three images are shown concurrently in each trial (Fig 1a) and remain on the screen until the participant makes their response. The top image (sample) depicts an object from one viewpoint. For the two images below, one depicts the same object as the sample from a different viewpoint (target) and the other depicts a different object from the same category (lure). Viewpoints for all objects were sampled randomly from a sphere around the object. The task for humans is to simply choose which of the target and lure images is the same object depicted in the sample image. Similar tasks have been used previously to probe for 3D shape representations in both models and humans \cite{rajalingham2018large, bonnen2021ventral}. Rajalingham et al (2018) showed that ImageNet CNNs are aligned to human shape matching judgements when the target and lure objects are drawn from different categories, but show a large gap to humans when the target and lure objects are both from the same category. This provides the motivation for our design where target and lure pairs are always drawn from the same category, so solving the task requires more fine-grained shape judgements rather than being solvable via categorization. For all experiments presented in this paper, we collect human behavioral responses (N=200) online using Prolific (www.prolific.co).

Stimuli were rendered from objects in the ShapeNet dataset \cite{shapenet2015}. We use a subset of ShapeNet that has $30565$ training objects and $8736$ test objects drawn from 13 manmade categories. Match-to-sample trials were generated by randomly pairing two test objects from the same category, then creating two trials with those objects' renders (one where object A is the sample/target and object B is lure, and one where object B is the sample/target and object A is the lure). Each participant only saw one trial for a given pair so all objects were novel throughout the experiment. Examples can be seen in Fig 1a.

\begin{figure}[h]
\begin{center}
\includegraphics[width=\columnwidth]{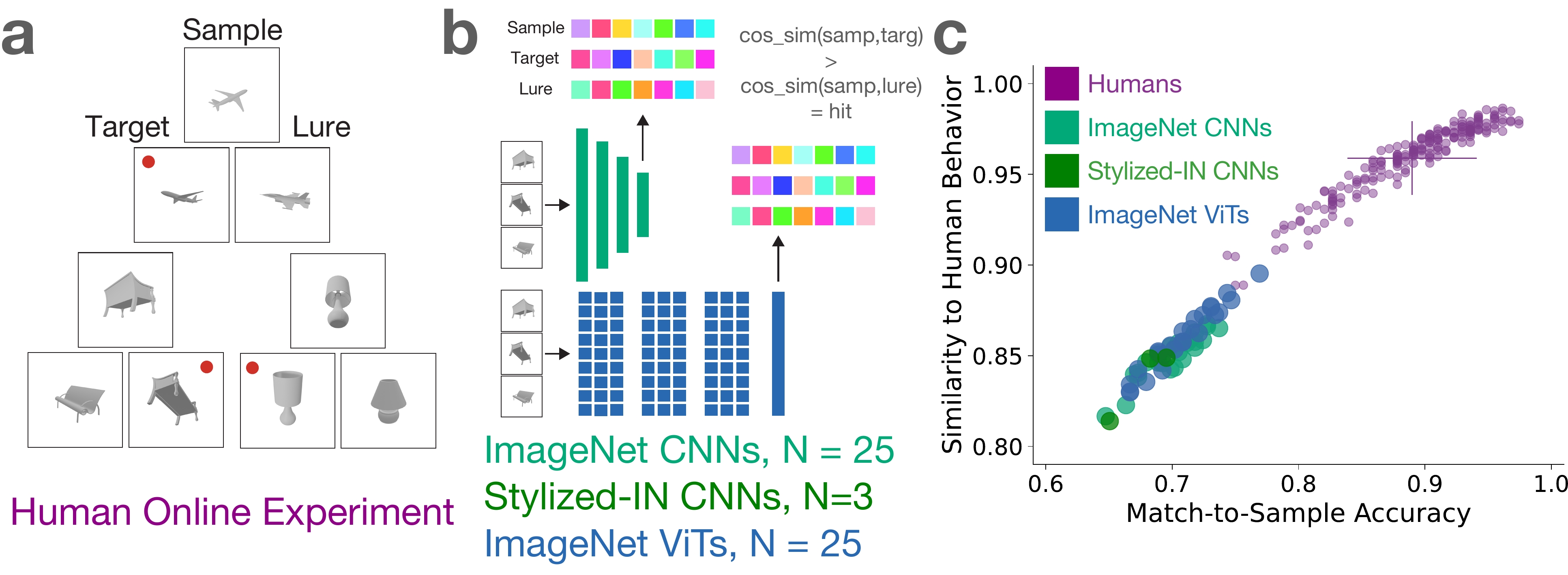}
\caption[Light Field Network]{a.) Example trials from the concurrent match-to-sample task used to probe 3D shape representations in humans and models. Participants see all three images concurrently, and their task is choose which of the bottom two objects (target and lure) depict the top object (sample) from a new viewpoint. For visualization purposes in this figure, the correct responses are indicated with a red dot. b.) Model procedure for match-to-sample tasks. c.) Match-to-sample performance for within-category 3D shape judgements in humans and ImageNet CNNs/ViTs. Notice the large gap between ImageNet DNNs and humans for both accuracy and trial-level similarity. Each purple dot is a single human participant. For human data, the y-axis encodes the average self-similarity of one held out participant to the mean of all other human participants (noise-ceiling). The errorbars along the y-axis show one standard deviation in accuracies across participants, and the errorbars along the y-axis show one standard deviation in the leave-one-subject-out noise-ceiling across participants.} 
\label{fig:Fig1}
\end{center}
\end{figure}

To complete the analogous task on DNNs (Fig 1b), each of the three images for a given trial are fed into a given network and unit activity was extracted from the penultimate layer (before the classification layer) for each image. The match-to-sample task is completed via a similarity-based measure; cosine similarity is computed between the sample/target and sample/lure model features, and the trial is counted as correct if the sample-target similarity is higher than the sample-lure similarity. For the standard DNN model zoo, we use 25 CNNs from PyTorch [28] and 25 ViTs from the timm package (timm.fast.ai), all pretrained for object classification on ImageNet. We also test 3 resnet50 CNNs trained on Stylized-ImageNet (Stylized-IN), a version of ImageNet which texture cues related to object category and increase human-alignment on an adversarial texture vs shape categorization task \cite{geirhos2018imagenet}. 

As expected, we observe a large gap in both accuracy and trial-level similarity between ImageNet DNNs and human behavior (Fig. 1c). Humans performed well on the task ($M=0.89, STD=0.051$, Fig. 1c, x-axis). A noise ceiling was computed as the cosine similarity between a given participant’s trial-wise performance and the mean trial-wise accuracies across all other participants, computed in a leave-one-participant-out fashion and averaged across participants ($M=0.96, STD=0.020$, Fig. 1c, y-axis). As with the human noise-ceiling, similarity to human behavior for each DNN model was computed as the cosine similarity between the vector of model responses across trials and the average human accuracy for each trial. We see much lower performance for ImageNet CNNs, with gaps in both accuracy $(M=0.70, STD_{toHumanAccuracy}=-3.75)$ and trial-level similarity to humans $(M=0.85, STD_{toHumanNoiseCeiling}=-5.47)$. ImageNet ViTs showed a similar gap compared to humans as CNNs in both accuracy $(M=0.71, STD_{toHumanAccuracy}=-3.58)$ and trial-level similarity to humans $(M=0.86, STD_{toHumanNoiseCeiling}=-4.99)$. The three CNNs trained on Stylized-ImageNet, which is designed to induce a shape bias, performed slightly worse than standard ImageNet CNNs on accuracy $(M=0.68, STD_{toHumanAccuracy}=-4.21)$ and trial-level similarity to humans $(M=0.84, STD_{toHumanNoiseCeiling}=-6.11)$. These findings highlight the 3D shape processing gap that has been reported in the literature between standard DNNs trained on large corpuses of natural images and humans that we aim to address with the following work.

\subsection{3D Light Field Networks and humans make similar 3D shape judgements}

The specific 3D neural field we use here is a 3D Light Field Network (3D-LFN) \cite{sitzmann2021lfns}. Light fields are an idealized representation that captures the color of rays of light passing through every possible position in a scene at every possible orientation. A 3D-LFN instantiates this idea in a neural network by mapping from coordinates defining a ray through a volume to the RGB value of that ray. The mapping is implemented as a multi-layer perceptron (MLP), and once trained the weights of the MLP define the 3D light field, and thus the 3D shape, for a given object. 

The flow of information through 3D-LFNs, during both training and inference, follows three steps (Fig 2a): 1.) Infer a set of shape latents (256D) from an RGB image, 2.) Map from shape latents to the neural field weights, 3.) Given a camera position, query rays from the neural field to render out an image. To infer shape latents from images (1), an RGB image is encoded using a CNN (resnet50) and 256D shape latents are linearly read out from the last convolutional layer. To map from latents to the neural field weights (2), a hypernetwork (a neural network trained to output the weights of another neural network) implemented using a series of two-layer MLPs with ReLU activation functions. The neural field itself is implemented as an eight-layer MLP with ReLU activation functions. The inputs to the field are coordinates defining a ray that passes through the space encoded by the field, and the output is the RGB value of the queried ray. To render the output image (3), a camera position and image plane are defined, and rays that pass from the camera through the image plane are sampled. The coordinates defining each ray are passed into the neural field, and the returned RGB value becomes the RGB value in the rendered image at the point where that ray intersects the image plane.

During training, the 3D-LFN is optimized end-to-end using a 3D multi-view learning objective. The input image depicts an object from one viewpoint, and the model’s objective is to render the same object from a different viewpoint defined by a camera matrix. The rendered image is compared to the ground-truth image of the object from the new viewpoint, and mean-squared-error between the rendered and ground-truth images provides the loss used to update network’s weights. The loss is backpropogated through the weights of the hypernetwork and CNN encoder. The CNN encoder is not pretrained, and its weights are entirely learned in tandem with the 3D-NF. During inference for test images, the latents and weights defining the 3D-LFN MLP are computed via a single feedforward pass through the model. Camera matrices can then be provided to render out the new object from the specified viewpoints, but the 3D-LFN latents and weights feature-spaces are computed without needing camera matrices.

\begin{figure}[h]
\begin{center}
\includegraphics[width=\columnwidth]{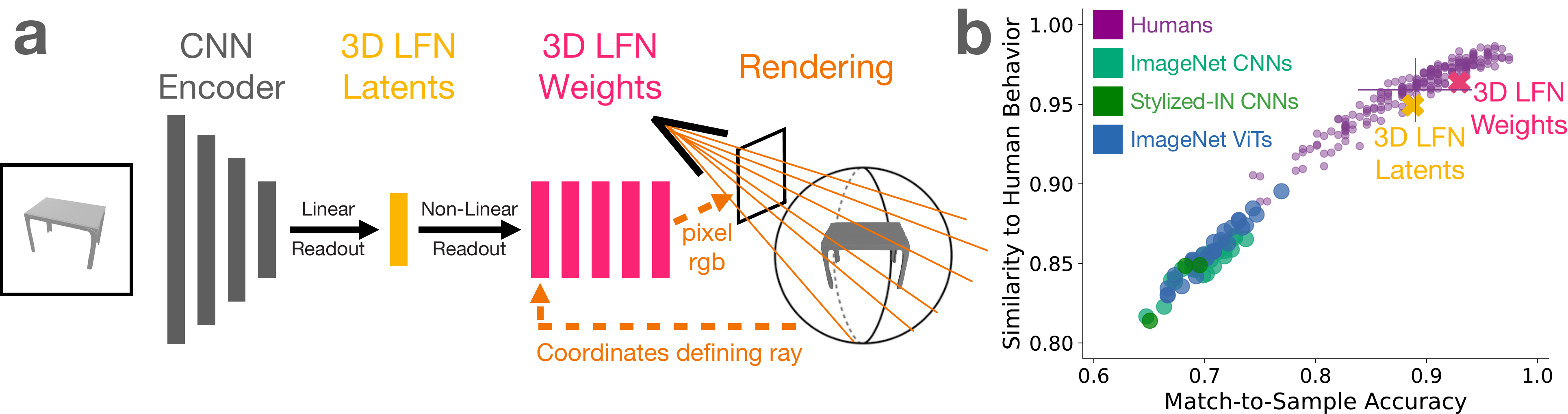}
\caption[Light Field Network]{a.) Schematic for the 3D-Light-Field-Network. At training time, the model is given viewpoint supervision and updated according to a 3D multi-view loss where the objective is to output the same input object from a new viewpoint. Features are extracted as the 3D LFN latents linearly read out from the CNN encoder and 3D LFN weights non-linearly read out from the latents during test inference, which is a single forward pass through the model. b.) Unlike ImageNet CNNs and ViTs, 3D-LFN features perform much better on the 3D matching task and approach human-level accuracy and trial-wise noise-ceiling. } 
\label{fig:Fig2}
\end{center}
\end{figure}

To synthesize the above, 3D-LFNs differ from standard DNNs (CNNs, ViTs) in 5 ways. We will revisit these components in a later section to elucidate what drives the 3D-LFNs ability to capture the 3D structure of objects. 1. While the encoder is a standard CNN resnet50, the overall architecture is different in that a series of MLPs, one set for the hypernetworks and another for the 3D field, are stacked on the encoder. 2. They are optimized on shape datasets in which images of objects are rendered from many viewpoints. 3. They are generative analysis-by-synthesis models that can produce reconstructed images of the represented objects. 4. They receive viewpoint supervision in the form of ground-truth camera matrices during training, which gives the model an explicit representation of 3D space. 5. They are trained with a 3D multi-view loss, in which the network’s learning objective is to render out the input object from a different viewpoint, which enforces a common 3D representation across viewpoints.

We trained a 3D-LFN on the ShapeNet dataset used for the match-to-sample task administered to humans, CNNs, and ViTs from the previous section. This training data includes 30565 objects from 13 object categories with 50 viewpoints per object, which are distinct from the test set of 8736 objects from which the stimuli for the match-to-sample experiments were rendered. The categories in the training and test sets are the same, but the specific objects are distinct. The shape latents and weights of the 3D-LFN are extracted as separate sets of features, and the task is performed using the same similarity-based measure that we applied to CNNs and ViTs. 

For the 3D-LFN latents (Fig. 2b), we see a marked improvement in both accuracy $(M=0.89, STD_{toHumanAccuracy}=-0.040)$ and trial-wise similarity to humans $(M=0.95, STD_{toHumanNoiseCeiling}=-.49)$. The 3D-LFN weights do even better (Fig. 2b), reaching the mean human accuracy $(M=0.93, STD_{toHumanAccuracy}=0.78)$ and within one standard deviation of the human noise ceiling $(M=0.96, STD_{toHumanNoiseCeiling}=0.24)$. Overall, these results show that the 3D-LFN supports 3D shape judgements that are aligned to humans for within-category comparisons, with both the latents and weights of the 3D-LFN falling within one standard deviation of both human accuracy and the human noise ceiling.

\subsection{3D-LFNs \& humans make similar 3D shape judgements for adversarial CNN-defined matching judgements}

Next, we used an adversarial stimulus-selection procedure to accentuate the failure modes of standard DNNs (i.e., where their performance deviates most sharply from humans). Using the 25 ImageNet CNNs, we filtered hundreds of thousands of potential within-category object pairs from ShapeNet to construct match-to-sample trials. We bin the object pairs according to the average CNN accuracy and sample from these bins to create a series of 5 adversarially-defined difficulty conditions. In the most difficult condition, the CNN average accuracy is $0.14$ (${STD_{toHumanAccuracy}=-6.95}$) and the trial-wise similarity to humans is $0.38$ (${STD_{toHumanNoiseCeiling}=-12.68}$). In the easiest condition, the CNN average accuracy is $0.68$ (${STD_{toHumanAccuracy}=-3.95}$) and the trial-wise similarity to humans is $0.83$ (${STD_{toHumanNoiseCeiling}=-5.81}$).

\begin{figure}[h]
\begin{center}
\includegraphics[width=\columnwidth]{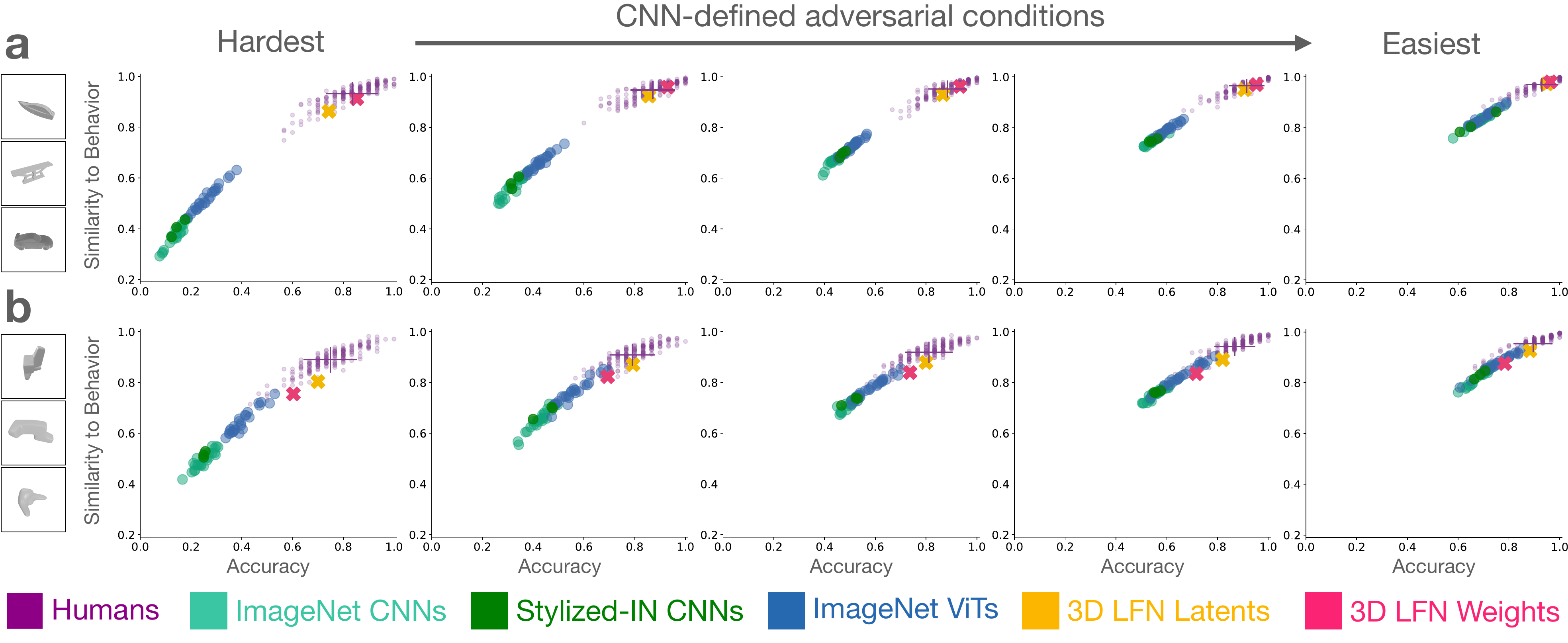}
\caption[Light Field Network]{Adversarial stimulus selection. a.) Many ShapeNet object pairs were filtered with the CNN model zoo to select match-to-sample trials that fall in five difficulty conditions that adversarially accentuate the difference between ImageNet trained models and humans. 3D-LFN latents and decoder features fall within the distribution of human responses for all conditions, including those where ImageNet models do poorly. b.) The same procedure was repeated for algorithmically generated objects with no category structure.} 
\label{fig:Fig3}
\end{center}
\end{figure}

Humans $({N=200})$ were largely unaffected by the CNN-defined difficulty conditions, showing high performance across all conditions (${M>0.83}$, Fig. 3a). ViTs (${STD_{toHumanNoiseCeiling}>-9.37}$) generally outperformed CNNs (${STD_{toHumanNoiseCeiling}>-12.68}$), but their performance was far from human alignment. Critically, 3D-LFNs were well-aligned with human behavior across difficulty levels for both latents (${STD_{toHumanNoiseCeilingLatents}>-1.58}$) and weights (${STD_{toHumanNoiseCeilingWeights}>-0.45}$). Across all difficulty conditions, the 3D-LFN weights were within 1 STD of the mean accuracy and noise ceiling compared to humans. These results show that even when selecting trials where standard CNNs and ViT models fail, 3D-LFNs support 3D shape inferences well aligned to humans.

Next, to ensure the previous results are not driven by the category-structure of the ShapeNet dataset, we ran the same experiment using procedurally-generated shapes with no explicit category structure. We repeated the full model training and adversarial match-to-sample task design pipeline using a set of shapes generated in Blender with ShapeGenerator (ShapeGen, Fig. 3b). There were 25k objects w/ 50 viewpoints per object used for training and 5k novel objects w/ 3 viewpoints per object for the set from which the same CNN-based selection procedure was applied. Viewpoints were sampled from a circle above each object with the camera pointing down 45 degrees at the object. A new 3D-LFN was trained on ShapeGen renders in the same fashion we used previously for ShapeNet.

As with ShapeNet, the humans (${N=200}$) show high accuracies $({M>0.75})$ regardless of the CNN-defined difficulty condition (Fig. 3b). In the most difficult condition, the 3D-LFN latents (${STD_{toHumanNoiseCeiling}=-1.78}$) and weights (${STD_{toHumanNoiseCeiling}=-2.77}$) are more aligned with human behavior than the ImageNet CNNs (${STD_{toHumanNoiseCeiling}=-8.09}$) and ViTs (${STD_{toHumanNoiseCeiling}=-4.98}$). The 3D-LFN latents is the most human-aligned model across the other four conditions (${STD_{toHumanNoiseCeiling}>-1.48}$), consistently beating out the CNNs (${STD_{toHumanNoiseCeiling}>-5.95}$) and ViTs (${STD_{toHumanNoiseCeiling}>-3.56}$). The performance of the ShapeGen-trained 3D-LFN is worse than for the ShapeNet-trained 3D-LFN, suggesting that the category structure of ShapeNet does increase the 3D-LFN's alignment to humans, but overall these results show that 3D-LFNs are more aligned to human 3D inferences than standard DNN models for objects that do not have an explicit category structure.

\subsection{What drives alignment between 3D-LFN \& human 3D shape judgements?}

To summarize thus far, we've shown that 3D-LFN models are well aligned to human 3D judgements compared to standard DNNs for within-category manmade object comparisons, across CNN-defined adversarial conditions for man-made objects, and across CNN-defined adversarial conditions for generated abstract objects. However, it is unclear which components of the 3D-LFN models are responsible for this alignment with humans. Fortunately, we can isolate the contribution of different components in these modeling frameworks using a series of computational experiments, which we outline below. 

The 3D-LFNs tested here differ from standard ImageNet CNNs and ViTs in five main ways. 1.) their architecture, which uses a standard CNN architecture as an encoder but stacks additional MLPs on top, 2.) they are trained on the rendered object datasets used for the human match-to-sample tasks rather than on a large corpus of natural images (e.g., Imagenet), 3.) 3D-LFNs are analysis-by-synthesis generative models that can visualize their learned latent space (i.e., reconstruct RGB images of the represented objects, given a novel viewpoint), 4.) 3D-LFNs receive viewpoint supervision during training; that is, ground-truth camera positions are provided to the 3D-NF to specify the viewpoint for the reconstructed image which gives the model an explicit conception of 3D space, and 5.) 3D-LFNs are trained with a 3D multi-view learning objective that enforces a shared representation across viewpoints. To determine which of these attributes is driving similar 3D shape judgments between humans and 3D-LFNs, we trained a series of new models to attempt to de-confound these attributes. 

To address the first two points, the 3D-LFNs have different architectural features than standard CNNs and ViTs and are trained on the ShapeNet/ShapeGen shape sets directly and not on a large corpus of natural images. In the following analyses, we will test two additional architectures, autoencoders and CNNs, to determine whether the architectural differences of 3D-LFNs are driving their alignment to humans. If the architectural differences of 3D-LFNs are the driving factor, we expect all subsequent models to perform poorly on the 3D shape matching tasks and alignment to human judgments compared to 3D-LFNs. Furthermore, all following models are trained using the same ShapeNet training set used to train the 3D-LFN. If simply training on rendered shape datasets is driving 3D-LFN performance and alignment, then all subsequent models should perform well on the tasks and show alignment to humans. 

First, we test whether having a generative component to render images from the latent space is driving human and 3D-LFN alignment. To answer this question, we train a standard 2D autoencoder that takes an image as input, maps it to a low-dimensional latent space using a resnet50 encoder, then reconstructs the same input image from the latent space using a series of deconvolutional layers (Fig. 4a). We find that the 2D autoencoder trained on ShapeNet performs within the distribution of 2D ImageNet CNNs (${M_{accuracy}=0.46}$, ${STD_{toCNNAccuracy}=1.13}$, ${M_{similarity}=0.67}$, ${STD_{toCNNSimilarityToHumans}=-0.054}$) and does not show alignment to humans (${STD_{toHumanAccuracy}=-7.52}$, ${STD_{toHumanNoiseCeiling}=-12.79}$) (Fig. 4e). Thus, simply having a generative component also does not seem to be sufficient to learn human-like 3D representations. Furthermore, the 2D autoencoder was trained on the same training images as the 3D-LFNs, so its poor performance indicates that simply training on rendered objects is not sufficient to learn human aligned 3D representations.

\begin{figure}[h]
\begin{center}
\includegraphics[width=\columnwidth]{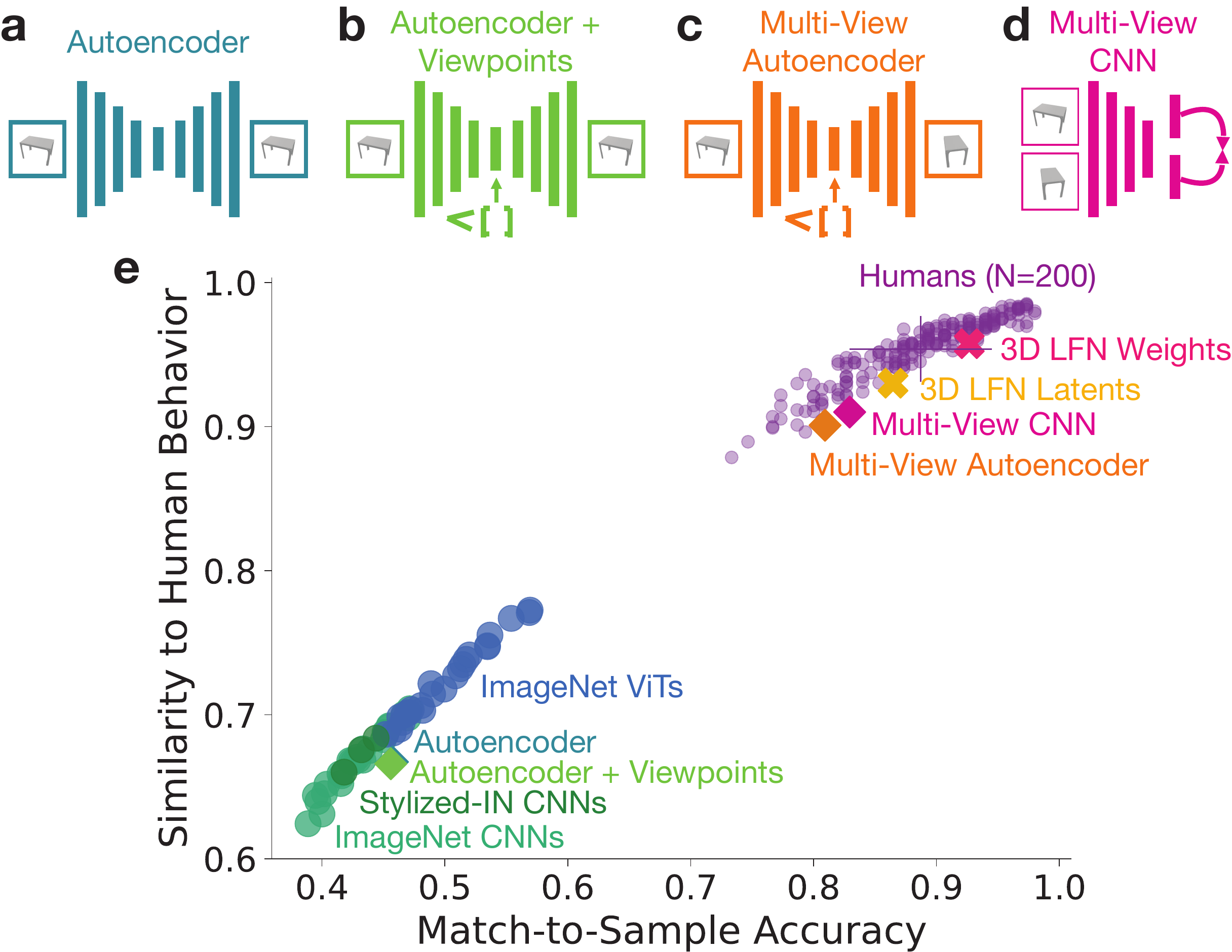}
\caption[Light Field Network]{Top row shows schematics for the various new models we train to determine which components support learning human-aligned 3D representations: a. Autoencoder, b. Autoencoder with viewpoint supervision, c. Multi-view autoencoder, d. Multi-view CNN. All models were trained end-to-end on the same ShapeNet dataset used to train the ShapeNet 3D-LFN b.) Results for all models on the adversarial CNN-guided ShapeNet match-to-sample trials averaged across difficulty levels. We find that a multi-view learning objective is the primary factor driving alignment to human 3D shape judgements.} 
\label{fig:Fig4}
\end{center}
\end{figure}

Next, we test whether receiving viewpoint supervision during training drives alignment to human 3D shape judgements. We use the same 2D autoencoder architecture, but now also append an embedding of the camera matrix defining the input/output viewpoint to the autoencoder latent space before it is fed into the decoder layers (Fig. 4b). However, we see that this does not improve performance over the vanilla 2D autoencoder, giving almost identical accuracy (${M=0.46}$, ${STD_{toHumanAccuracy}=-7.54}$) and trial-wise similarity (${M=0.67}$, ${STD_{toHumanNoiseCeiling}=-12.85}$) to humans as the autoencoder without viewpoint supervision (Fig. 4e). Thus, viewpoint supervision also appears not to be the main factor in learning 3D representations that support human-level 3D judgements. 

Finally, we investigate the role of the multi-view learning objective in driving alignment to human 3D shape judgements. By multi-view learning objective, we mean a learning objective that enforces similarity in representation across different viewpoints of the same object. For the 3D-LFNs, this means the model receives an image of an object from one viewpoint, and must render it from a different viewpoint. To incorporate something similar into an autoencoder, we take the 2D autoencoder with viewpoint supervision and make it a multi-view autoencoder by having the output viewpoint differ from the input viewpoint, as in 3D-LFNs (Fig. 4c). The camera matrix embedding appended to the autoencoder latent space defines the output viewpoint the model should render (again similar to 3D-LFN training). Here, we see a marked improvement in both 3D judgement accuracy (${M=0.81}$, ${STD_{toHumanAccuracy}=-1.36}$) and trial-wise similarity to humans (${M=0.90}$, ${STD_{toHumanNoiseCeiling}=-2.36}$) (Fig. 4e). While the multi-view autoencoder does not reach the human noise ceiling and is outperformed by the 3D-LFN latents and weights, it is a clear improvement over autoencoders without a multi-view learning objective, ImageNet CNNs, and ImageNet ViTs.

Pushing this line of argument further, both 3D-LFNs and the multi-view autoencoder described above incorporate a multi-view learning objective, but also each have a generative component and receive viewpoint supervision. To test whether a multi-view learning objective is sufficient by itself to enforce learning human-like 3D representations, we train a final multi-view similarity CNN using a resnet50 architecture (Fig. 4d). The model uses a modified MOCO contrastive loss \cite{he2020momentum}. In standard MOCO contrastive CNNs, different image manipulations (e.g. translation, left-right flip, negative colors) are applied, and the model's loss is designed to make the embeddings computed from two modified versions of the same image as similar as possible relative to the embeddings computed for other images. We use this same scheme, but rather than applying image manipulations we provide two different viewpoints of the same ShapeNet object. The model still has a multi-view learning objective in that it must associate different viewpoints of the same objects as similar in its learned embedding space, but without any generative capability or viewpoint supervision. Indeed, we find that this multi-view similarity CNN also does quite well on the 3D match-to-sample tasks (Fig. 4e), outperforming the multi-view autoencoder in accuracy (${M=0.83}$, ${STD_{toHumanAccuracy}=-1.01}$) and trial-wise similarity to humans (${M=0.91}$, ${STD_{toHumanNoiseCeiling}=-1.95}$) but still falling short of 3D-LFNs, human accuracy, and the human noise ceiling. Nonetheless, by simply training a standard CNN architecture to have similar embeddings across viewpoints of the same object, we can learn a 3D shape representation that makes much more human-aligned 3D shape judgements than standard ImageNet CNNs and ViTs.

Overall, these computational experiments show that training on rendered shape datasets, having generative capabilities to render images of a model’s latent space, and viewpoint supervision providing an explicit conception of 3D space are all insufficient to produce human-aligned 3D shape representations. The primary factor that drives alignment to human 3D shape judgements in our experiments is training models with exposure to multiple viewpoints of the same objects and a multi-view learning objective that enforces a common 3D representation that generalizes across viewpoints. Architecture is still relevant: the 3D LFNs are more aligned to human behavior than standard DNN architectures modified with a multi-view loss, but the right training data (exposure to multiple viewpoints of objects) and the learning objective (enforcing a common representation across viewpoints) play a larger role.

\subsection{3D-LFNs and multi-view DNNs are not human-aligned for novel shape categories outside their training set}

Finally, we test the extent to which the 3D-LFN and multi-view DNN models generalize to novel shape categories not included in training. Human 3D vision is able to represent novel shapes we have never encountered in our visual experience. If we come upon an abstract sculpture in a park, we are able to perceive its shape even if it occupies a new part of our shape space that has never been activated. In these experiments, humans' ability to generalize to novel shapes is evident in the high human accuracies for the match-to-sample task with abstract ShapeGen objects (Fig. 3b). One possibility is that the 3D-LFN and multi-view DNNs learn a similarly broad shape space that can generalize to novel categories of objects they have not seen before. The alternative possibility is that the 3D-LFN and multi-view DNNs learn a space that can generalize to novel exemplars of categories included during training (as we have shown in all previous experiments) but fail for new categories of shapes that fall outside their visual experience.

To test the generality of the shape spaces learned by the 3D-LFN, multi-view autoencoder, and multi-view similarity CNN, we re-trained new versions of each model on ShapeNet holding out all cars, planes, and chairs in turn. We then constructed 3D match-to-sample experiments from only objects in each held-out category using the same adversarial stimulus selection procedure as earlier. With the CNN model zoo, we filter many possible object pairs for each held-out category and select the 150 pairs that best exemplify the performance gap between humans and standard DNN models. As in previous experiments, we collected human responses for each held-out category’s new match-to-sample trials online (N=210, N per category=70).

\begin{figure}[h]
\begin{center}
\includegraphics[width=\columnwidth]{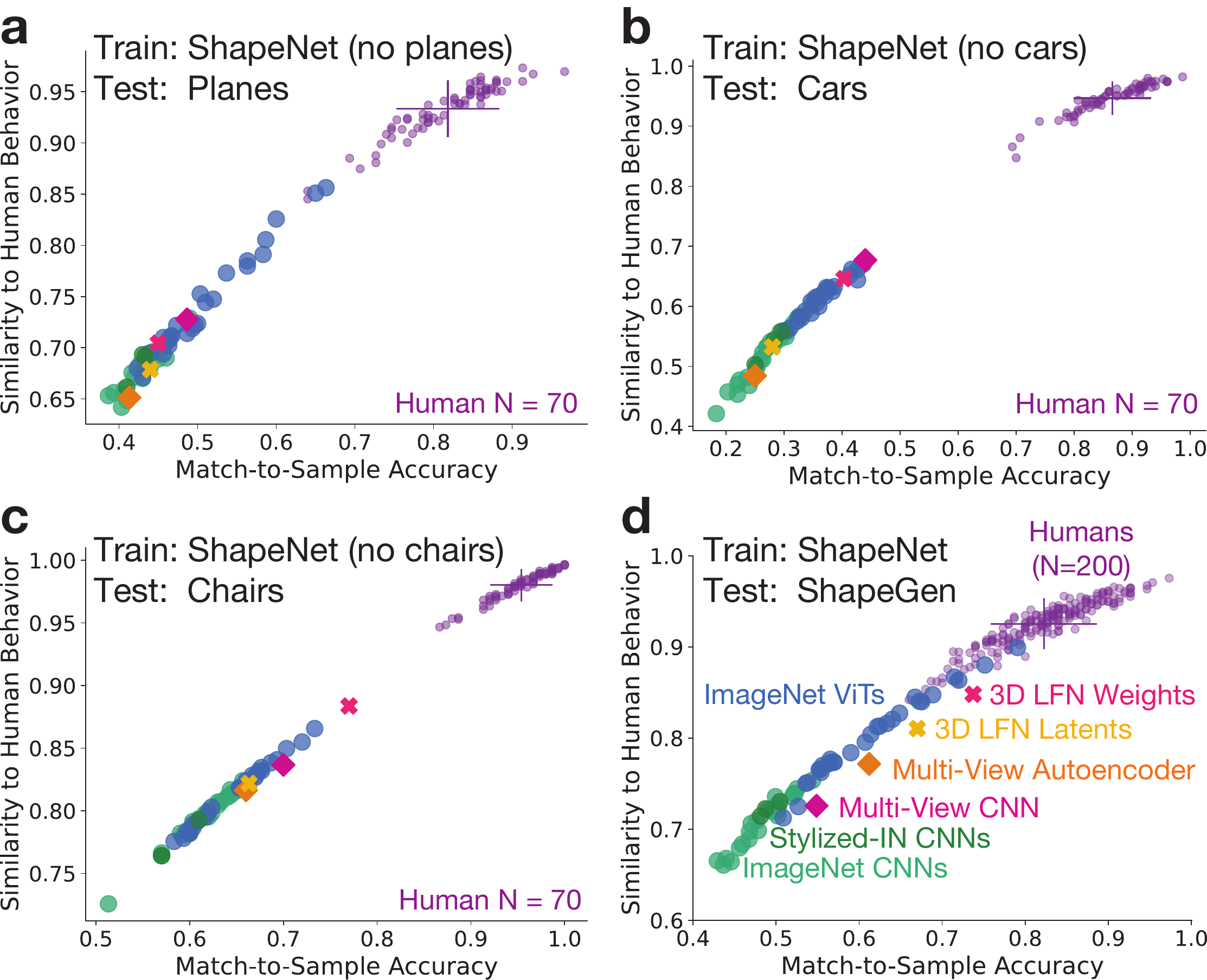}
\caption[Light Field Network]{Out-of-distribution generalization for 3D models to novel object categories. For each category (a. planes, b. cars, c. chairs), all objects from that category were held out during training for the 3D-LFN and multi-view DNN models. Match-to-sample trials were adversarially selected for the held-out using the CNN model zoo to accentuate the performance gap between ImageNet DNN models and humans. We observe that the 3D-LDN and multi-view DNN models struggle to generalize to novel object categories not included in their training distribution, falling within the range of standard CNN and ViT models and well below human performance. However, for cars and chairs the best performing models were still models trained with a multi-view learning objective. d. The 3D-LFN and multi-view DNNs from previous experiments trained on all ShapeNet categories were tested on the ShapeGen match-to-sample experiment consisting of abstract generated objects. Again, the models fall below human performance and noise-ceiling when generalizing out of their training distribution.} 
\label{fig:Fig5}
\end{center}
\end{figure}

For each new task, we find that the 3D-LFN and multi-view DNNs fail to reach human-level performance for held-out object categories not included in their training set. For planes (Fig. 5a), the best performing multi-view DNN is the multi-view similarity CNN (${STD_{toHumanNoiseCeiling}=-7.62}$), but ImageNet ViT models (${STD_{toHumanNoiseCeiling}=-7.03}$) on average outperform the 3D-LFN and multi-view DNNs. For cars (Fig. 5b), the best performing multi-view model is again the multi-view similarity CNN (${STD_{toHumanNoiseCeiling}=-10.11}$), but despite being the best performing model overall it still falls far short of human alignment. For chairs (Fig. 5c), the 3D-LFN weights are most aligned to humans (${STD_{toHumanNoiseCeiling}=-7.87}$), but again the improvements over standard DNN models are small and fall well below human performance and noise ceiling.

Next, we tested whether the 3D-LFN and multi-view DNNs trained on the full set of ShapeNet categories generalize to the ShapeGen match-to-sample trials consisting of abstract objects. We take the same ShapeNet-trained 3D-LFN, multi-view autoencoder, and multi-view similarity CNN from the previous section and tested them on the ShapeGen match-to-sample experiment, averaging performance across the 5 adversarially-defined difficulty conditions. Again, we observe all models performing below the human performance and noise ceiling (Fig. 5d). The best multi-view DNN is the 3D-LFN weights (${STD_{toHumanNoiseCeiling}=-2.85}$). The 3D-LFN and multi-view DNNs achieved similar accuracies as the ImageNet DNNs, but with slightly lower trial-wise similarity to humans for standard DNNs with equivalent accuracies. The fact that ShapeNet-trained models perform somewhat better when tested on ShapeGen (Fig. 5d) than on ShapeNet categories they were not exposed to during training (Fig. 5a-c) most likely reflects the smaller viewpoint variations entailed for the ShapeGen judgements (where views are sampled from a circle above the object with the camera pointing down 45 degrees) than the ShapeNet judgements (where views are sampled randomly from a full sphere around each object).

Together, these results show that the 3D-LFN and multi-view DNN models tested here do not learn a generalizable shape space that supports human-aligned 3D inferences for novel shape categories not included in the training diet. The 3D LFN and multi-view DNNs did not fail completely to generalize; out of the four experiments, the 3D LFN weights or multi-view similarity CNN not trained on the object categories used for the match-to-sample trials were still the most human-aligned models in two cases, with the other 3D and multi-view models performing within the distribution of ImageNet DNNs. However, model-human fell short of prior experiments where the test categories used to construct the match-to-sample tasks were included in the models’ training diet. These results highlight an important limitation of the 3D-LFN and multi-view DNN models presented here to be addressed by future research.

\section{Discussion}

Here we address one of the key outstanding gaps between the visual abilities of neurally mappable DNNs and humans: inferring the 3D shape of objects. To address this computational challenge, we start with a relatively novel class of DNNs, 3D neural fields, that have received little attention to date from the cognitive (neuro)science community. We show that a 3D-LFN model supports human-aligned 3D inferences across three human psychophysics experiments, identify the key requirement for achieving alignment with human 3D shape judgements, and highlight the failure of these 3D models to fully generalize to novel shape categories outside their training distribution. Critically, we find that a common 3D representation across viewpoints must be enforced in DNNs to capture human performance, and that the 3D-LFNs and multi-view DNNs presented here fail to generalize to novel shape categories outside their training distribution. These computational findings highlight the potential for studying human 3D vision in a neurally mappable computational framework as well as the limitations of current 3D and multi-view DNNs, providing a roadmap for building more human-like 3D vision models moving forward.

What are the necessary conditions for a DNN model to make human-aligned 3D shape judgements? We trained a series of autoencoder and CNN models on the same rendered shape datasets as the 3D-LFN incorporating generative abilities, viewpoint supervision, and 3D learning objectives, and find that similarity between humans and models on 3D matching tasks is driven by exposure to multiple viewpoints of the same objects during training and a 3D multi-view learning objective. We show that at least two different types of 3D learning objectives produce models with behavior similar to humans. For 3D-LFNs and multi-view autoencoders, 3D learning is encouraged with a multi-view analysis-by-synthesis learning objective (given input image from viewpoint A, render same object from viewpoint B). For the multi-view similarity CNN, we directly update learned embeddings from two viewpoints of an object to be as similar as possible. For ShapeNet manmade objects, the 3D-LFN best matched human accuracies and trial-wise patterns of behavior, but all models trained with a 3D multi-view objective, including a multi-view autoencoder and a multi-view similarity CNN, showed much greater alignment to humans than a variety of DNNs trained without a multi-view objective. These findings suggest that there may be several types of learning objectives that support human-aligned 3D shape judgements by enforcing a common 3D representation for multiple viewpoints of the same objects. That is, it is not that standard architectures cannot learn humanlike shape representations; they just need to be trained on the right data and task.

The importance of seeing objects from multiple viewpoints to support 3D visual learning, found here for computational models, is in line with research characterizing the visual diet of human infants and toddlers using head-mounted cameras in natural contexts \cite{smith2018developing}. Frontward views of faces are the most common visual input in infant’s first 3 months, a pattern that continues but decreases over the following 9 months as seeing body parts of other humans and hands manipulating objects, especially during mealtimes, become increasingly common \cite{jayaraman2015faces, fausey2016faces, jayaraman2017faces, clerkin2017real}. From ages 1-2, the presence of hands, particularly hands manipulating objects, becomes the most frequent input in the visual diet of toddlers \cite{fausey2016faces}. Crucially, the viewpoints at which human toddlers see objects via their own self manipulation does not appear to be random, but curated to support visual statistical learning. 18-24 month old toddlers manipulating objects are biased towards planar views of the objects, and the proportion of spontaneous planar views predicts subsequent 3D recognition performance on those objects \cite{james2014young}. Furthermore, while part of the bias towards planar views is driven by ease of holding the object, a preference for planar views persists when controlling for ease of holding, suggesting those views provide useful visual features and surface properties for learning the 3D shape of objects \cite{james2014some}. Finally, even for 4-7 month old infants, ability to recognize the 3D shape of objects is most strongly linked to visual-motor manipulation abilities which allow infants to expose themselves to many views of a novel object \cite{soska2010systems}. Considering this empirical evidence, it is perhaps less surprising that computational models trained on standard datasets (e.g., Imagenet) do not spontaneously learn to represent objects in 3D, given there is no direct training on multiple viewpoints of the same object. Similarly, these empirical data corroborate our computational results: models that do get such multi-view exposure support human-aligned 3D shape judgements. 

Recent work on learning online invariances in CNNs also supports the importance of multi-view exposure during training to learn 3D representations. Biscione \& Bowers (2022) showed that CNNs can develop online invariances to common transformations, but only when the model’s training set included variation along the relevant dimension \cite{biscione2022learning}. They trained CNNs for object classification or same-different judgements on images rendered from ShapeNet with variation in the viewpoint, rotation, scale, translation, and brightness of the objects. Evaluating on a similar multi-view match-to-sample task as the one used here, they find that CNNs trained on images with a particular transformation learn invariance to that transformation. Most similar to the current work, they find that CNNs can learn to be viewpoint-invariant, but only when exposed to multiple viewpoints and rotations of the same object during training. While they do not evaluate the alignment between their models and humans, they suggest that special architectures may not be necessary to learn human-like invariances for viewpoint, position, and lighting provided the right training diet and task. These predictions are consistent with our findings that a variety of DNN architectures are aligned to human 3D shape judgements, but only when the learning objective enforces learning a common representation across two viewpoints of the same object. While a special architecture directly representing 3D space was most aligned to human behavior (3D-LFNs), a standard CNN architecture with a contrastive loss enforcing common representation across viewpoints showed a marked jump in alignment to human 3D shape judgements. Future work will further adjudicate the role of architectures, special or otherwise, in achieving human-like 3D shape judgements, but exposure to multiple-viewpoints of objects during training is evidently essential.

We aimed to solve the challenge of training DNNs with human-aligned 3D representations and identifying what supports such learning within 3D and multi-view DNNs. Our work succeeded in addressing a primary gap between the visual ability of humans and DNNs. However, this does not mean the current results are evidence that the 3D models presented here are necessarily what humans are doing to compute the 3D shape of objects. Several aspects of our models are clearly not biologically plausible. Next, we will outline these shortcomings, focusing on the graphics techniques used to support analysis-by-synthesis in 3D-LFNs and the learning procedures used to train the models, which highlight the next challenges for modeling human 3D shape perception in a more biologically plausible fashion.

3D-LFNs are DNN analysis-by-synthesis models that regress 3D shape latents from a CNN, map the latents to the 3D field space, then use a series of graphics techniques to render an image of the represented object from a viewpoint specified by a provided ground-truth camera matrix. The graphics techniques—casting rays through an image-plane to render an RGB image of the depicted object—are implemented using standard graphics tools that have to date not been implemented in artificial neural networks or shown to be implemented by biological neurons. These techniques are one engineering solution to compute the 3D multi-view loss in an analysis-by-synthesis model. We are not proposing that the brain is using computer graphics techniques such as rendering and casting rays (or backpropogation for that matter) to learn 3D representations; this is simply one avenue to enforce 3D representations in models to produce human-aligned 3D shape judgements. An important direction for future research is to identify novel learning objectives that support 3D learning, perhaps via analysis-by-synthesis techniques, that can be implemented on neurons without needing standard graphics tools.

Another biologically implausible aspect of the current models is the training dataset and procedures. All 3D models here are trained on artificial datasets in which object meshes are rendered from many random viewpoints, and pairs of images depicting the same object from different viewpoints are provided on individual training trials. The training procedure is a form of supervised learning, in which objects must be rendered from many viewpoints, then models explicitly directed via a 3D multi-view learning objective to learn a shared representation for objects that generalizes across viewpoints. Additionally, for the 3D-LFN and multi-view autoencoder models, a ground-truth camera matrix is provided during training to specify the output viewpoint from which to render the image, another form of direct 3D supervision. Furthermore, as with most DNNs, the training dataset requires many examples, of both objects and viewpoints for each object, to then generalize at test time to novel objects not included in the training set. Humans, on the other hand, do not receive curated datasets depicting many thousands of objects from many views or ground-truth viewpoint positions to define an explicit 3D space. Instead, they receive such information as they watch objects move and rotate and as they themselves move about the world \cite{smith2018developing}. Furthermore, infant and toddler headcam data show that their visual diet consists of many views of a small number of objects, with sparser views for a larger set of objects \cite{clerkin2017real, smith2018developing}. A key challenge for 3D models moving forward will be developing learning algorithms that scaffold 3D representations from unsupervised video input without camera poses, in line with human infants and toddlers learning 3D from ego-motion, that do not require training sets of tens of thousands of objects to achieve generalizable 3D representations. Solving such challenges in the next generation of 3D models is also likely to help address the failures to generalize to novel object categories not included in the training diet that we observe for the 3D DNNs presented here.

One of the primary strengths of DNN models of vision is the neural mappability of their internal representations, which has allowed for the prediction of visually-evoked brain responses (e.g. electophysiology, fMRI) from images \cite{yamins2014performance, khaligh2014deep}. Where in the brain might we expect 3D DNN models, such as the 3D-LFN presented here, to predict neural responses to objects?  Decades of work has been used to suggest that the late stages of processing with the ventral visual stream (VVS) represent object shape \cite{dicarlo2012does}. However, more recent work suggests that the VVS may, in fact, simply provide a basis space of texture-like representations which can be used to infer shape-level object properties \cite{jagadeesh2022texture}. This claim inverts a common critique of standard DNNs (e.g., \cite{geirhos2018imagenet}), suggesting that the VVS, like CNNs, is ‘texture-biased.’ More broadly, it’s possible that while some 3D shape properties might be represented within the VVS, there are neuroanatomical structures beyond the VVS which are critical for this ability. In recent work, Bonnen et al. (2021) has shown that medial temporal cortex (MTC), downstream from the VVS, plays a causal role in 3D object perception. Remarkably, MTC-lesioned human performance is approximated by Imagenet-trained CNNs, while MTC-intact performance dramatically exceeds model performance—causally implicating MTC in computing the 3D shape of objects. We note that these MTC-dependent visual inferences also have a characteristic behavioral profile—integrating over visuospatial sequences—which is not captured in any computational models of 3D perception. These data provide clear neuroanatomical and algorithmic constraints on models which aim for more human-like 3D object perception. The structure of the 3D-LFN and 3D autoencoder models, in which 3D shape is regressed from a CNN using a series of additional computational modules bears obvious similarity to the idea that VVS captures a textural basis space that feeds into downstream regions like MTL that recover 3D shape. Additionally, the dorsal stream has been implicated in 3D shape processing \cite{sereno2002three, van2016posterior}, suggesting that the sorts of structured 3D representations captured by 3D DNNs may share capture more variance in dorsal neural responses than standard DNNs. 

In summary, we have shown that 3D-LFN models approximate some aspects of human 3D visual inferences. Furthermore, this work highlights the importance of exposure to the same objects from multiple viewpoints for DNNs to get anywhere close to human-like 3D matching behavior. In doing so, our work addresses one of the central critiques over the application of DNNs for modeling human vision. Of course, there are obvious differences between the current 3D models and biological brains, highlighted by our 3D models’ failure to generalize to novel object categories not included during training. Nonetheless, resolving these engineering challenges is a critical step in demonstrating that DNNs are capable of human-aligned 3D inference. We leave it to future work to use neural and behavioral constraints, DNN approaches, and other modeling frameworks, such as probabilistic programming, to design and evaluate more biologically plausible computational models of human 3D object perception. 

\section{Methods}

\textbf{3D shape datasets.} Two 3D shape datasets were used in these experiments: ShapeNet \cite{shapenet2015} and ShapeGen (blendermarket.com/products/shape-generator).

ShapeNet is a large collection of 3D object meshes drawn from a variety of manmade categories. We use the 13 largest ShapeNet categories, with a total of $39301$ objects across all categories. The categories are: airplanes, tables, cars, lamps, chairs, stereos, benches, phones, guns, couches, cabinets, boats, monitors. $30565$ objects were assigned to a training set for learning 3D models, and $8736$ were assigned to the test set. All objects used in the match-to-sample tasks for humans and models were drawn from the test set for all experiments. Using Blender \cite{blender}, we rendered 50 images of each object from viewpoints randomly sampled from a sphere around the object. The objects were rendered matte grey without color or textures to emphasize shape processing in the models and experiments. 

ShapeGen is a tool for procedurally generating abstract objects in Blender. It provides a simple interface through which a user can generate abstract objects by fusing together several simpler shapes of different sizes, each of which can be rotated, beveled, smoothed, or extruded according to predefined user specifications. To generate the shapes in our dataset, we predefined ranges for several parameters and sampled values uniformly within those ranges, which provided smooth and interesting variation across shapes. More concretely, our shapes were generated by taking a single cube as a base object (whose initial length and width is set by a random seed), extruding a random face on that object between 5-10 times (each time at a random angle, and random length), and rounding the edges by applying a catmull-clark modifier to the resulting mesh, to create a smooth shape. 25000 objects were generated for the training set, and 5000 objects were generated for the test set. As with ShapeNet, all objects used in the match-to-sample tasks were drawn from the test set, Blender was used to render 50 viewpoints of each object, and objects were rendered matte grey. The viewpoints for ShapeGen objects were sampled from a circle above each object with the camera pointing 45 degrees down towards the object.

\textbf{Human behavioral experiments.} Human behavioral experiments were conducted online using Prolific. Participants were each paid \$15/hr for participating.
Experiments were approved by the MIT Committee on the Use of Humans as Experimental Subjects. Participants were notified of their rights
before the experiment, and were free to terminate participation at
any time by closing the browser window. Experiments consisted of an initial set of instructions, 6 practice trials with feedback, and 150 main trials with no feedback. Participants were not able to progress to the main experiment until they successfully completed all 6 practice trials. Each experiment took an average of 15 minutes for participants to complete. Experiments were presented to participants using a standard JsPsych toolkit \cite{jspsych}. Participants were screened for participation in prior studies in this paper, so each participant only appears once and in just one experiment.

For all experiments, trials were constructed by assigning two objects to a pair. For each pair of objects, two trials were constructed, one with object A as the sample and target and B as the lure, and one vice versa. In the behavioral experiments, participants were split into two batches and each batch only saw one trial for a given object pair ensuring that all objects were novel for each trial.

\textbf{Model Zoo of Standard DNNs.} The model zoo of standard DNNs consisted of 25 pretrained convolutional neural networks (CNNs) and 25 pretrained visual transformers (ViTs). 

The 25 CNNs were pretrained for object classification on ImageNet and downloaded from PyTorch (pytorch.org). The models used were: alexnet, densenet121, densenet161, densenet169, densenet201, resnet101, resnet101wide, resnet152, resnet18, resnet34, resnet50, resnet50wide, resnext101, resnext50, shufflenetv2, squeezenet1.0, squeezenet1.1, vgg11, vgg11bn, vgg13, vgg13bn, vgg16, vgg16bn, vgg19, vgg19bn. For all analyses, the unit activity for the penultimate layer before the classification layer were extracted.
  
The 25 ViTs were pretrained on ImageNet and downloaded from the timm (timm.fast.ai) package. As with the CNNs, all models were pretrained for object classification on ImageNet. The models used were: convit\_base, convit\_small, convit\_tiny, mvitv2\_base, mvitv2\_base\_cls, mvitv2\_huge\_cls, mvitv2\_large, mvitv2\_large\_cls, mvitv2\_small, vit\_base\_patch16\_224, vit\_base\_patch16\_clip\_224, vit\_base\_patch32\_clip\_224, vit\_base\_patch8\_224, vit\_base\_r50\_s16\_224, vit\_large\_patch14\_clip\_224, vit\_large\_patch32\_224, vit\_relpos\_base\_patch16\_224, vit\_relpos\_base\_patch16\_clsgap\_224, vit\_relpos\_medium\_patch16\_224, vit\_relpos\_medium\_patch16\_rpn\_224, vit\_small\_patch16\_224, vit\_small\_r26\_s32\_224, vit\_srelpos\_medium\_patch16\_224, vit\_srelpos\_small\_patch16\_224, vit\_tiny\_r\_s16\_p8\_224. For all analyses, the unit activity for the penultimate layer before the classification layer were extracted.

\textbf{3D Light Field Networks.} The essence of a neural field generally is a neural network model that takes some set of coordinates as input and outputs a property at those coordinates. Common examples of fields are magnetic or gravitational fields, both of which encode the continuous function of magnetic or gravitational force across space. In computer graphics, 3D shape fields are learned by mapping from coordinates in space (e.g. xyz, ray coordinates) to color or volume properties at that location. Once the neural network is trained, an image depicting the encoded space from a given viewpoint can be rendered by querying the neural field to recover shape/color. The most commonly used neural fields, neural radiance fields (NeRF) \cite{mildenhall2021nerf}, take xyz coordinates as input and output RGB and volume density at the queried coordinate. The field is then queried using a volumetric renderer to produce images of the scene from a given viewpoint. NeRFS have been used in computer graphics to create photorealistic encodings of objects and scenes by overfitting the field with many (>100) views from an individual scene. More relevant for modeling perception, a class of neural fields called conditional neural fields compute the continuous 3D function for objects from images (conditional because the field is conditioned on the input image). The models used here fall into this latter camp.

3D Light Field Networks (3D-LFNs) \cite{sitzmann2021lfns} are models that compute a continuous function defining the light field of a volume in space using neural networks.  3D-LFNs follow three stages, each with their own architecture: 1. Inferring 3D shape latents from images using a CNN encoder, 2. Predicting the weights of the neural light field from the shape latents using hypernetworks, and 3. Using the neural light field to render an image from a given camera viewpoint. These models are part of a broader class of models in computer vision called conditional neural fields \cite{xie2022neural}, which estimate neural fields from images rather than fit them directly to a given scene with many viewpoints, as in common in computer graphics \cite{xie2022neural}. 

The inference stage (1) of the 3D-LFNs used here was implemented as a resnet50 CNN encoder. The final classification layers were removed, and a linear layer was used to map from the final convolutional layer to a $256$D fully-connected layer that represents the 3D latent space for the model. To map from shape latents to the weights of the 3D light field (2), we used a hypernetwork. A hypernetwork is a class of neural networks used in metalearning \cite{ha2016hypernetworks} that is trained to predict the weights of another neural network. The hypernetwork is implemented as a series of two layer multi-layer perceptron (MLP) with $256$ units per layer that takes the $256$D shape latents as inputs and outputs the weights for one layer of the neural field. There is one hypernetwork for each adjacent pair of layers in the neural field. The neural light field itself is implemented as an 8 layer MLP. The first layer input is a six dimensions for the pl\"ucker coordinate defining a ray through the scene, the output layer is three dimensions for the RGB value of the ray, and the hidden layers have $256$ hidden units. For the rendering procedure during training, a ground-truth camera matrix is provided, an image-plane defined at a fixed distance between the camera and object, and rays are queried such that they pass through the camera plane from the camera position. One ray is queried per pixel in the output render, and the RGB value returned by the neural field for a given ray becomes the RGB value for the pixel where that ray intersects the image plane. The loss for training the model is the mean-squared-error between the output render and the ground-truth image of the object from the same camera position. The loss is backpropogated to update the weights in the hypernetworks and CNN encoder.

For all 3D-LFN analyses, we extracted the shape latents and weights defining the 3D-LFN MLP as features. The 3D-LFN latents are a 256D fuly-connected layer linearlly read out from the CNN encoder. The 3D-LFN weights were dimensionality reduced to a 150D feature-space using PCA before subsequent analysis.

\textbf{Autoencoders.} We implemented 3 different types of autoencoders. The first is a standard autoencoder that takes an image as an input, runs it through a resnet50 encoder and maps to an 256D latent space, then uses a series of deconvolutional layers to map from the latent space back to reconstruct the input image. In our experiments, we use this as a standard CNN-based architecture that contains a generative component as in 3D LFNs but without viewpoint supervision or a multi-view learning objective.

Next, we use the same autoencoder architecture but make one modification so it receives a form of viewpoint supervision. We take the camera matrix that defines the input and output viewpoint of the rendered object, embed it using the same embedding scheme as NeRF \cite{mildenhall2021nerf} to a 256D vector, append the embedded camera matrix to the latent space of the autoencoder, then pass both the latents and embedded camera pose together as the input to the decoder deconvolution layers. This now produces a model that has both a generative component and viewpoint supervision, as in 3D LFNs, but without a multi-view learning objective. 

The final autoencoder we use is identical to the above version with viewpoint supervision, but modified to support a multi-view learning objective. In this version, the input image and output rendering target are two different viewpoints of the same object. The camera matrix embedding appended to the latent space now defines the target output viewpoint the deconvolutional block should reconstruct. Now, the model has a generative component, viewpoint supervision, and a multi-view learning objective like the 3D LFN. The goal was to make a model that had as much as possible in common with the 3D LFN, just with a different architecture and rendering procedure. All of these models were implemented with custom PyTorch code.

For all autoencoder models, the match-to-sample tasks were completed using the same similarity-based measure as all other models applied to the ND latent space.

\textbf{Multi-View Similarity CNN.} For the multi-view similarity CNN, we used a standard resnet50 CNN architecture with a modified Momentum Contrast (MOCO) style learning objective \cite{he2020momentum}. In standard MOCO models, a target image is augmented in two ways (possible manipulations: random cropping, random cropping + resizing, color jittering, random greyscale conversion, random horizontal flip, random gaussian blur). Each augmented image is passed through the same resnet50 and mapped to an 2046D output latent space. The model’s learning objective is to make the latents for the two ablated versions of the input image as similar as possible (positive samples) relative to augmented versions of other images in the same minibatch (negative samples).

To modify this for a multi-view learning objective, we instead provide the model with two images of the same object from different viewpoints (without any augmentations). The model’s objective then becomes to make the two output latents for different viewpoints of the same object as similar as possible relative to negative samples from un-augmented images of other objects in the minibatch. We used a modified version of the implementation in the PyContrast repository (github.com/HobbitLong/PyContrast). 

\textbf{Model alignment to human behavior.} Human and model behaivor was compared in a trial-level fashion using cosine similarity. Average trial-level human accuracies were computed across participants for each trial, producing a vector of accuracies. For models, the binary correct/incorrect performance for each trial was also arranged into a vector over trials. Similarity to Human Behavior was computed as one minus the cosine distance between the human and model accuracy vectors. 

The human noise-ceiling was computed using a similar procedure within the human data. In a leave one out fashion, the similarity between the binary correct/incorrect vector for one participant and the average accuracies for the remaining participants was computed. This is done for each participant, and the average across participants is the noise-ceiling for the model comparisons.

\section*{Data Availability}

The software packages and datasets used to create all stimuli are available online:
\begin{itemize}
 \item ShapeNet.v2 Dataset (shapenet.org)
 \item Blender (blender.org)
 \item ShapeGenerator Blender Plugin (blendermarket.com/products/shape-generator)
\end{itemize}

The following Python packages were used for the modeling:
\begin{itemize}
 \item All analyses were completed using Python 3.6 (python.org)
 \item Code to train Light Field Networks is available on GitHub (github.com/vsitzmann/light-field-networks)
 \item The ImageNet CNNs are available in Pytorch (pytorch.org) 
 \item The ImageNet ViTs are available in timm (timm.fast.ai)
 \item The Multi-View Similarity CNN was trained with a modified version of PyContrast (github.com/HobbitLong/PyContrast)
 \item The Autoencoders were implemented with custom code in PyTorch, which will be made available when the paper is peer-reviewed and accepted for publication
\end{itemize}

Human behavioral data and custom Python code used for the model-human alignment experiments will be made available when the paper is peer-reviewed and accepted for publication.

\section*{Contributions}
TPO, JBT, and NK conceived of the research. TPO, NK, and TB designed the experiments with input from JBT. YF and TPO collected the human behavioral data. TPO constructed the models with assistance from AT and VS. TPO conducted the modeling and behavioral analyses with input from TB. TPO, NK, TB, and JBT interpreted the results. TPO, TB, and NK wrote the paper with input from JBT, YF, and VS.

\section*{Acknowledgments}
This work was supported by National Institutes of Health grant DP1HD091947 awarded to Nancy Kanwisher, Office of Naval Research MURI grant PO \#BB01540322 awarded in part to Josh Tenenbaum, and the Center for Brains, Minds, and Machines (CBMM) funded by NSF STC award CCF-1231216. We thank Kevin Smith, Ratan Murty, Tucker Smith, and Shiloh Kanwisher for helpful feedback on this work.

\bibliographystyle{unsrt}  
\bibliography{references}  

\end{document}